%% file: main.tex
\documentclass[10pt, conference, compsocconf]{IEEEtran}
\IEEEoverridecommandlockouts    
\usepackage[utf8]{inputenc}
\usepackage{amsmath,amssymb,amsfonts}
\usepackage{bm}
\usepackage{url}
\usepackage{graphicx,graphics}
\usepackage{multirow}
\usepackage{array,booktabs}
\usepackage{multirow}
\usepackage[ruled,linesnumbered]{algorithm2e}
\usepackage{textcomp}
\usepackage{tabularx}
\usepackage{relsize}
\usepackage[flushleft]{threeparttable}
\usepackage{cite}
\usepackage[table,svgnames]{xcolor}
\usepackage{paralist}
\usepackage[export]{adjustbox}
\usepackage{soul}
\usepackage{arydshln}
\usepackage[normalem]{ulem}
\let\labelindent\relax
\usepackage{enumitem}
\newcommand{\midsepremove}{\aboverulesep = 0.2mm \belowrulesep = 0.2mm}
    \midsepremove
\newcommand{\midsepdefault}{\aboverulesep = 0.605mm \belowrulesep = 0.984mm}
    \midsepdefault

\newcommand{\ayush}[1]{\textcolor{blue}{#1}}

\soulregister \cite7
\soulregister \ayush7
\setstcolor{Red}

\begin{document}\sloppy
\bstctlcite{IEEEexample:BSTcontrol}
%
%
%
\title{\LARGE \bf
Local Region-to-Region Mapping-based Approach to \\Classify Articulated Objects
}
\author{\IEEEauthorblockN{Ayush Aggarwal$^{*}$, Rustam Stolkin, Naresh Marturi}
\IEEEauthorblockA{Extreme Robotics Laboratory,\\
School of Metallurgy and Materials, University of Birmingham (UoB)\\
Birmingham, United Kingdom\\
Email: $^{*}${\tt axa1508@student.bham.ac.uk}, {\tt r.stolkin@bham.ac.uk}, {\tt n.marturi@bham.ac.uk} 
\thanks{This work was supported by the UK National Centre for Nuclear Robotics (NCNR). Part funded by CHIST-ERA under Project EP/S032428/1 PeGRoGAM and in part supported by Faraday Institution sponsored Recycling of Lithium Ion Batteries (ReLiB) project (grant: FIRG005).}}
}
%
\maketitle
%
%
\input{sections/abstract}
%
\section{Introduction}
\label{sec:introduction}
\input{sections/introduction}

\section{Proposed Method}
\label{sec:method}
\input{sections/method}

%
\section{Experimental Validations}
\label{sec:experiments}
\input{sections/experiments}
%
\section{Conclusion}
\label{sec:conclusion}
\input{sections/conclusion}


\end{document}

%% file: sections/abstract.tex
\begin{abstract}
Autonomous robots operating in real-world environments encounter a variety of objects that can be both rigid and articulated in nature. Having knowledge of these specific object properties not only helps in designing appropriate manipulation strategies but also aids in developing reliable tracking and pose estimation techniques for many robotic and vision applications. In this context, this paper presents a registration-based local region-to-region mapping approach to classify an object as either articulated or rigid. Using the point clouds of the intended object, the proposed method performs classification by estimating unique local transformations between point clouds over the observed sequence of movements of the object. The significant advantage of the proposed method is that it is a constraint-free approach that can classify any articulated object and is not limited to a specific type of articulation. Additionally, it is a model-free approach with no learning components, which means it can classify whether an object is articulated without requiring any object models or labelled data. We analyze the performance of the proposed method on two publicly available benchmark datasets with a combination of articulated and rigid objects. It is observed that the proposed method can classify articulated and rigid objects with good accuracy.
\end{abstract}

\begin{IEEEkeywords}
Object Classification, Articulated Objects, Articulated Classification
\end{IEEEkeywords}

%% file: sections/introduction.tex

\begin{figure*}
\centering
\includegraphics[width=0.9\textwidth]{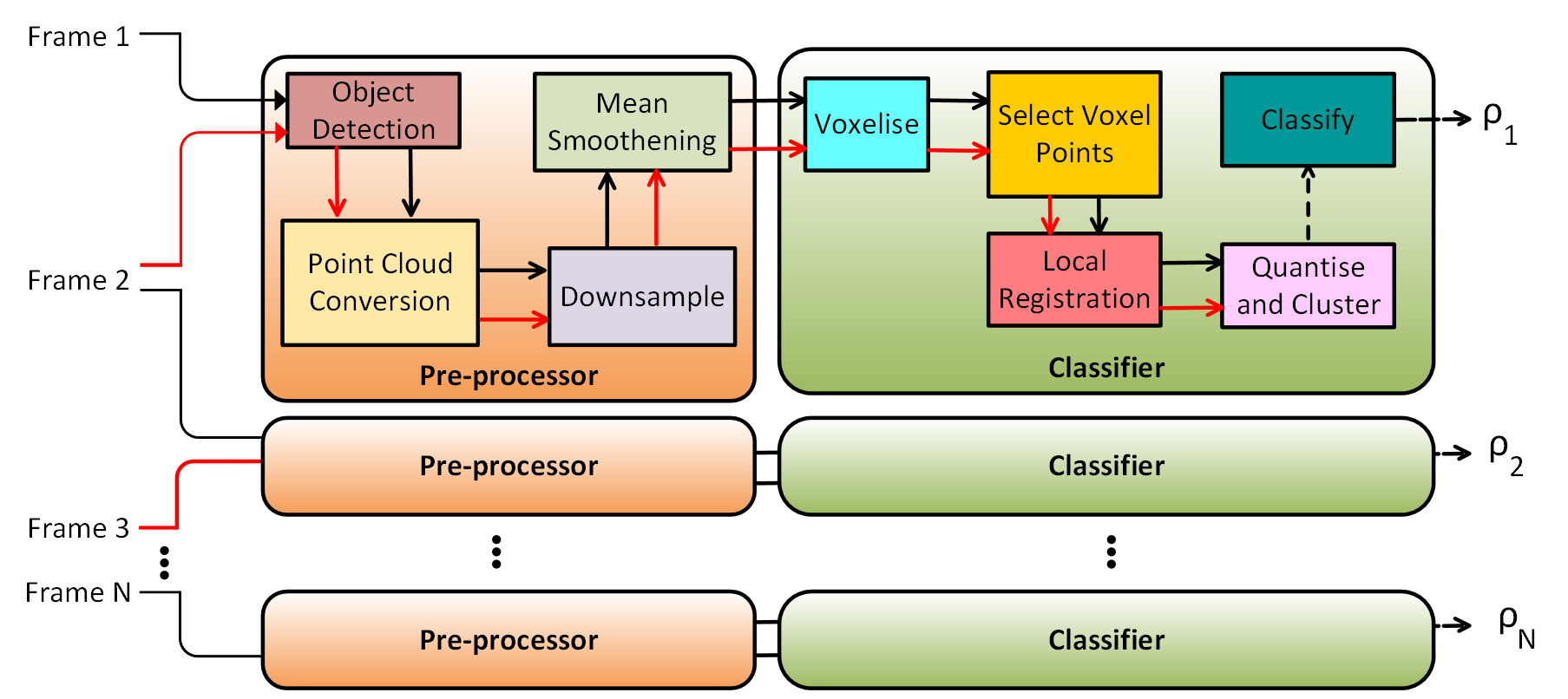}
\caption{The pipeline of the proposed registration-based object classification method. For each set of frames $i, i+k \in N$, where $k$ is frame skip, we first perform pre-processing on the depth map ${\bf D}$ and RGB image ${\bf C}$ for all the frames to obtain the point cloud ${\bf P}$ of the object from the scene. These clouds are then downsampled to obtain ${\bf P}^{d}$. Both the point clouds \textit{i.e.} ${\bf P}_{i}^{d}$ and ${\bf P}_{i+k}^{d}$ are then voxelised to obtain ${\bf V}$ voxels. For each voxel grid ${\bf V}_{m}$, points in the voxel region are selected in both ${\bf P}_{i}^{d}$ and ${\bf P}_{i+k}^{d}$, which is then used to perform registration to obtain a local transformation ${\bf T}_{m}$. All the obtained local transformations are then quantised and clustered to obtain the unique set of local transformations in hash table $\mathcal{H}$. Based on this set, a decision $\rho_i$ on this set is made to belong to one of the classes. Finally, we utilise a moving max-count filter on the decisions vector $\boldsymbol{\rho}$ to classify the object.}
\label{fig:pipeline}
\end{figure*}

Several regularly used household and industrial objects are uniquely represented as articulated, i.e., objects composed of multiple rigid links that are kinematically linked, e.g., doors, chains, clamps etc. Identifying such types of objects in general scenes (also filled with other rigid objects) is of prime interest for many robotic applications. Specifically, knowledge of object properties like rigidity and articulation (in addition to geometric visual features) helps in reliable tracking and pose estimation for robotic and vision applications, e.g., robot tasked with clearing unknown objects in case of hazardous decommissioning tasks \cite{marturi2016towards}. 
It is highly challenging for a robot to operate in the scenes with articulated objects without knowing these specific object properties. Motivated from this, we propose a classification technique to detect if an object being manipulated is either rigid or articulated.

As mentioned, articulated objects are composed of multiple rigid structures. When manipulated, each of these structures move differently relative to the type of the joint they are linked with. These movements are classified into three categories: revolute, prismatic and free-form, which also help in kinematic modelling. %
In the literature, modelling, pose estimation, and tracking of articulated objects has been studied extensively \cite{TowardsUAO, markerless_motion_capture, articulated_free_form_hand, Understanding_3d_articulation_in_internet_video, visual_articulated_parts_identification, model_articulated_in_image, online_recursive_perception,  sar_images, articulated_tracking_render_parts,articulated_object_recognition_hough_transform}. In~\cite{TowardsUAO}, authors presented a method to learn a kinematic model of an articulated object from a given video sequence. In this method, they first learn the possible object articulation model and then verify the learned model on the testing sequence. The method is designed to handle free form movements. However, it is constrained to objects with only two rigid parts. Furthermore, the method relied on markers for accurate object position, which may not be available in real scenarios. 
A marker-less object skeleton estimation method from multi-view point cloud was proposed in~\cite{markerless_motion_capture}, wherein the authors utilise the generated skeleton curvatures to align and form the model of the object. Although the method is marker-less, it requires a complete point cloud of the object. In comparison, the proposed method can work on single-view (marker-less) point clouds from a video sequence to classify the articulated objects.

Several deep learning-based methods for estimating object articulation type and articulation axis are proposed in the literature~\cite{articulated_free_form_hand, Understanding_3d_articulation_in_internet_video, visual_articulated_parts_identification}. In~\cite{articulated_free_form_hand}, the authors proposed an approach that predicts object and hand models to understand object-hand interactions. Articulated object detection method using RGB-D video sequence was proposed in~\cite{Understanding_3d_articulation_in_internet_video}, wherein the authors first detect the object parts plane and axis over each image and then utilise the temporal relation between the frames to predict the object bounding box, articulation planes and axis. In~\cite{visual_articulated_parts_identification}, the authors used object RGB-D images and corresponding part segmentations to predict the kinematic constraint between the parts. These learning-based methods exhibit good performance and robustness towards noise; however, they require accurately labelled training data, which may not be always available in practical real-world scenarios. Furthermore, with free-form articulated objects, there are numerous possible object states, making labelling a time-consuming and extensive process. To overcome this limitation, we propose an online classification method that can directly infer the object's type without the need for labelled data. In~\cite{model_articulated_in_image}, the authors proposed a part segmentation-based model creation and tracking system on images, where they create shape-based models for each rigid component of the object and track them to form a complete articulation model. Finally, in~\cite{online_recursive_perception}, a three-step articulation state estimation method is proposed by processing RGB-D images over an interaction sequence. 

The aforementioned methods are designed to estimate the model and kinematics of articulated objects. However, they all assume that the scene object is articulated. We believe it is important to first analyse whether the object in the scene is articulated before attempting to model or track it. To achieve this, we perform a temporal local region-to-region registration on 3D point clouds obtained from an observed sequence of object movements/interactions. Using the video corresponding to the object's movements, we first generate a mask for the object of interest in each video frame using an off-the-shelf object localization method. Then, using this generated mask and the corresponding depth information, a point cloud of the object is created for each frame. For each set of consecutive frame point clouds, we perform local region registrations, which result in local transformations from one frame to the other. These local transformations are then quantized (to reduce the effect of cloud noise) and clustered to obtain the final set of unique transformations between the local regions of the two frames. For a rigid object, these transformations are the same for all local regions, while for articulated objects, the local transformations may not be the same, and the local regions of different articulated parts may move with different rotations or translations. %
\begin{figure*}
\centering
\includegraphics[width=0.95\textwidth]{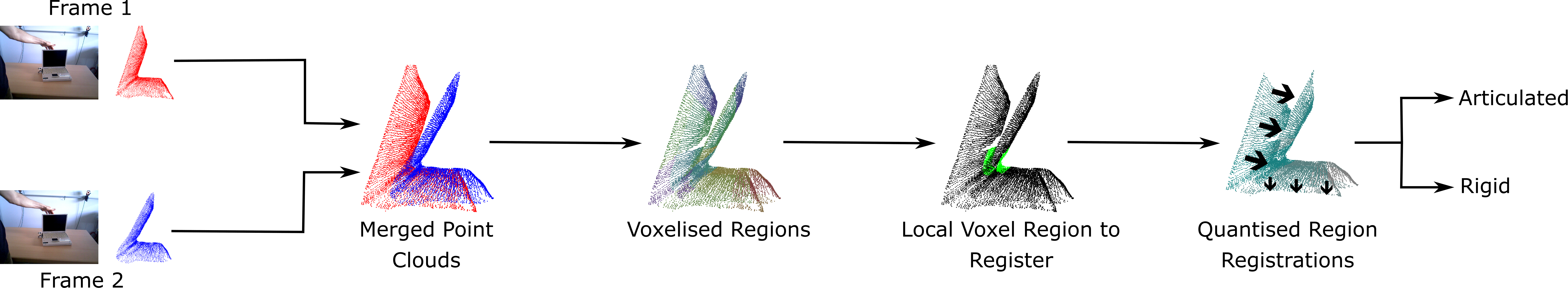}
\caption{The in-frame classification process. Given the pre-processed object point cloud from two frames (e.g. Frame 1 and Frame 2), these clouds are at first merged and then split into voxel grids (marked with coloured regions) to object the local regions from both clouds. Each voxel grid is then selected (marked with green colour) for local registration. The registrations are then quantised and the regions with similar registration are clubbed together (as shown in last image with arrows marking the magnitude and direction of transformation). Finally, after counting the number of unique registration, the motion of object in these two frames is classified as articulated or rigid.}
\label{fig:pipeline_disp}
\end{figure*}
%
%
The key contributions of this paper are summarised as follows:
\begin{itemize}
    \item We propose a new model-free object classification method to identify if an object is rigid or articulated by observing a sequence of object movements.
    \item We design a constraint-free registration-based approach with local region-to-region mapping to detect object with any type of articulation during classification.
\end{itemize}

The main advantage of the proposed method is that it is a constraint-free approach, which can classify any type of articulated object, without being limited to specific types of articulations. The proposed method is a parametric approach and does not require any training, making it independent of any labelled data requirements. We analysed the performance of the proposed method using two publicly available benchmark datasets containing a mixture of rigid and articulated objects. The results obtained demonstrate the ability of our approach to classify articulated objects.

%
%

%% file: sections/method.tex
In this section, we describe the proposed framework for articulated object classification. 
For demonstration, we consider a $N$-frame video of an object being manipulated where each frame consists of a depth map and an rgb image. Then for each set of consecutive frames ${i, i+k} \in N$, we perform registration between local regions of corresponding point clouds in order to classify the motion between them as no motion, rigid motion, or articulated motion. It is worth noting that the point cloud of a frame is generated using the depth information of intended object regions. This process is presented in detail in Sec. \ref{subsec:preprocess}. To perform this task, initially, each frame is passed through a set of pre-processing steps for noise and background filtering. Afterwards, the corresponding filtered point clouds of the frames are passed through the classifier. The complete pipeline of our approach is shown in Fig.~\ref{fig:pipeline}. Majorly, the pipeline is divided into two modules, i.e., the pre-processor and the classifier. Each of these modules are described in the following subsections.

\subsection{Pre-Processing} \label{subsec:preprocess}
In general scenario, the depth map ${\bf D}_{i}$ and rgb image ${\bf C}_i$ of a given frame $i$ may consist of a variety of items other than the object of interest, e.g. wall, table, etc. For the proof of concept, we assume that in any sequence of frames only a single object is being manipulated and all the remaining objects are stationary. Hence, all these other objects can be considered as unwanted noise elements which impact the performance of the classifier. Further, the presence of these objects creates a logical problem in accurate classification, which is discussed in the next subsection. 

\begin{algorithm}
\caption{Pre-Processing of input frame depth map and rgb image}
\label{alg:pre-processing}
\SetAlgoLined
\SetKwFunction{ObjectDetection}{ObjectDetection}
\SetKwFunction{Crop}{Crop}
\SetKwFunction{ConvertToCloud}{ConvertToCloud}
\SetKwFunction{OutlierNoiseRemoval}{OutlierNoiseRemoval}
\SetKwFunction{VoxelDownsample}{VoxelDownsample}
\SetKwFunction{MeanSmoothening}{MeanSmoothening}
\KwData{{Depth map ${\bf D}_i$}, rgb image 
${{\bf C}_i}$ for $i$th frame, object id $O$, Camera intrinsic ${\bf c}_{K}^{i}$, outlier noise standard deviation $s$, voxel downsample size $v$, smoothening radius $r$}
\KwResult{Cropped Point Cloud ${\bf P}_{i}^{d}$}
$p, bb =$ \ObjectDetection{${{\bf C}_i}, O$}\;
\eIf{$p \neq None$}
{
${\bf D}_{i}^{c} =$ \Crop{${\bf D}_{i}, {bb}$}\;
${\bf P}_{i} =$ \ConvertToCloud{${\bf D}_{i}^{c}, {\bf c}_{K}^{i}$}\;
${\bf P}_{i}^{o} =$ \OutlierNoiseRemoval{${\bf P}_{i}, s$}\;
${\bf P}_{i}^{v} =$ \VoxelDownsample{${\bf P}_{i}^{o}, v$}\;
${\bf P}_{i}^{d} =$ \MeanSmoothening{${\bf P}_{i}^{v}, r$}\;
}
{Skip Frame}
\end{algorithm}

In this pre-processing stage, we process each frame to remove the unwanted elements and noise from the point cloud. At first, we perform object localisation in the rgb image \textbf{${\bf C}_i$} of the frame using the well-known Mask-RCNNv2 \cite{mask_rcnn_v2} mask generation model. We utilise the pre-trained weights over Microsoft COCO dataset~\cite{coco} and do not perform any fine-tuning on the considered objects. From the Mask-RCNNv2, we obtain a list of detection score and mask ($bb$) (no classifications utilised) over the objects in the frame image. From these predictions, we create mask instances per object. To select the intended object, we identify the mask instance, between two consecutive frames, that is being manipulated. If an object is being manipulated, we consider the respective mask. However, if no object is found in the frame, we skip that frame and move to process the next set of frames. Considering the case when the intended object is detected in ${\bf C}_i$, we crop out the object and remove rest of the information from depth map to obtain ${\bf D}_{i}^{c}$.

This process is performed for all the frames. Once the intended object region is obtained, we convert the object depth map to 3D point cloud ${\bf P}_{i}$ using the depth camera intrinsic parameters ${\bf c}_{K}^{i}$. ${\bf P}_{i}$ is further processed to remove the outliers and noise using statistical noise removal, to obtain ${\bf P}_{i}^{o}$. This processed point cloud is then downsampled using a voxel size $v$. This downsampled point cloud ${\bf P}_{i}^{v}$ is then passed through a mean smoothness filter with radius $r$ to remove the sensor noise. The smoothened point cloud ${\bf P}_{i}^{d}$ is then passed to the classifier module. Pre-processing steps are summarised in Algorithm~\ref{alg:pre-processing}. Note that any other object localisation method~\cite{saliency_survey,sam,freesolo,segmentation_without_label} can also beused instead of Mask-RCNNv2 to obtain the object mask. Further, the learning methods are only utilised for mask generation, and are not part of the proposed algorithm.

\subsection{Classifier}
The corresponding filtered point clouds of the pre-processed frames ${i}$ and ${i+k}$ are used by this module to classify if they belong to any of the following three classes:
\begin{description}
    \item[i) No Motion (NM)] -- when there is no motion in object between the frames $i$ and $i+k$.
    \item[ii) Rigid Motion (RM)] -- when there is a rigid motion in the object between the frames $i$ and $i+k$.
    \item[iii) Articulated Motion (AM)] -- when there are multiple unique motions in the object between the frames $i$ and $i+k$.
\end{description}

This classification is performed for all the frame sets in the given video, henceforth called as \textit{\bfseries in-frame classification}. The proposed in-frame classification method is summarised in Algorithm~\ref{alg:classifier}. In this method, we first split ${\bf P}_{i}^{d}$ and ${\bf P}_{i+k}^{d}$  into voxel grids ${\bf V}$ with the voxel size $x$. This gives us $M$ voxel grids with points in them. Then, for each voxel grid ${\bf V}_m, m \in M$, we consider the points from ${\bf P}_{i}^{d}$ and ${\bf P}_{i+k}^{d}$ in voxel grid ${\bf V}_m$ as ${\bf P}_{i}^{d, m}$ and ${\bf P}_{i+k}^{d, m}$ respectively. With the assumption that the object motion between two frames is not very large, we perform Iterative Closest Point (ICP)~\cite{icp} based registration between ${\bf P}_{i+k}^{d, m}$ and ${\bf P}_{i}^{d, m}$. As it is assumed that the points of both the voxels are close, the initial transformation for ICP is provided as identity matrix ${\bf I}$. This registration results in a homogeneous transformation matrix ${\bf T}_{m}$ with rotation ${\bf R}_{m}$ and translation ${\bf t}_{m}$ components for the $m$th voxel grid. The rotation matrix is converted to quaternion form ${\bf q}_{m}$ for further processing. 

\begin{algorithm}
\caption{Articulated and rigid object classification using point clouds}
\label{alg:classifier}
\SetAlgoLined
\SetKwFunction{ICPRegistration}{ICPRegistration}
\SetKwFunction{Voxelise}{Voxelise}
\SetKwFunction{SelectVoxelPoints}{SelectVoxelPoints}
\SetKwFunction{Euclid}{Euclid}
\SetKwFunction{ToQuaternion}{ToQuaternion}
\KwData{source point cloud ${\bf P}_{i}^{d}$, target point cloud ${\bf P}_{i+k}^{d}$, voxel size $x$, quaternion quantisation $\overline{q}$, translation quantisation $\overline{t}$, moving window size $w$}
\KwResult{In-frame object class $\rho_{i}$}
${\bf V} =$ \Voxelise{${\bf P}_{i}^{d}, {\bf P}_{i+k}^{d}, x$}\;
$M = |{\bf V}| $\;
$m = 0$\;
$\mathcal{H} = \{\}$\;
\SetKwRepeat{Do}{do}{while}
\Do{$m \leq M$}
{
\tcp{Local voxel region selection}
${\bf P}_{i}^{d,m} =$ \SelectVoxelPoints{${\bf P}_{i}^{d}$}\;
${\bf P}_{i+k}^{d,m} =$ \SelectVoxelPoints{${\bf P}_{i+k}^{d}$}\;
${\bf T}_{m} = $ \ICPRegistration{${\bf P}_{i}^{d,m}, {\bf P}_{i+k}^{d,m}$}\;
${\bf R}_{m} = {\bf T}_{m}[:3, :3]$\;
${\bf t}_{m} = {\bf T}_{m}[:3, 3]$\;
${\bf q}_{m} = $ \ToQuaternion{${\bf R}_{m}$}\;
\tcp{Quantisation operation}
$\Tilde{\bf q}_{m} = \lfloor\frac{{\bf q}_{m}}{ \overline{q}}\rfloor * \overline{q}$\;
$\Tilde{\bf t}_{m} = \lfloor\frac{{\bf t}_{m}}{ \overline{t}}\rfloor * \overline{t}$\;
\tcp{Hash table is generated with unique keys by clustering}
\eIf{$[ \Tilde{\bf q}_{m} | \Tilde{\bf t}_{m}] \in \mathcal{H}$}
{
$\mathcal{H}[\Tilde{\bf q}_{m} | \Tilde{\bf t}_{m}] = [\mathcal{H}[\Tilde{\bf q}_{m} | \Tilde{\bf t}_{m}];{\bf V}_{m}]$\;
}
{$\mathcal{H}[\Tilde{\bf q}_{m} | \Tilde{\bf t}_{m}] = [{\bf V}_{m}]$\;
}
$m = m + 1$\;
}
$K = |\mathcal{H}|$\;
\uIf{$K \geq 2$}
{
$\rho_{i} = AM$\;
}
\uElseIf{$K = 1$}
{
$\rho_{i} = RM$\;
}
\Else
{
$\rho_{i} = NM$\;
}
\end{algorithm}

For each $m \in M$, ${\bf q}_m$ and ${\bf t}_{m}$ are quantised with $\overline{q}$ and $\overline{t}$ to remove any noise errors in the registration process.These quantised values are then joined and hashed into a hash table $\mathcal{H}$, wherein for each key, we store the respective voxel grids. Due to the possible articulated nature of the objects, it is observed that some portion of an object may become occluded or go out of the frame within consecutive frames (e.g. a box opening may occlude the top portion of the box). To overcome this issue, in the proposed method, we skip the voxel grids which do not have any correspondence match during ICP. Further, if a match has low confidence score, we also skip those registrations. After these skips, if more than $\alpha\%$ of the voxel grids are matched successfully, we progress with the classification step, otherwise those frames are considered to be non-reliant and the process moves to the next set of frames.

Utilising the prepared hash table $\mathcal{H}$, we perform in-frame classification based on the following rules: \begin{itemize}
    \item if number of keys in $\mathcal{H} > 1$, then the motion is articulated and classified as AM;
    \item if the number of keys in $\mathcal{H} = 1$, and the key corresponds to a rotation / translation, the motion is RM;
    \item if the number of keys in $\mathcal{H} = 1$ and the key represents no motion, i.e., $q \in 0^{1\times4}$ and $t \in 0^{1\times3}$, then it is NM.
\end{itemize}  
The in-frame classification method is shown in Fig.~\ref{fig:pipeline_disp} and is summarised in Algorithm~\ref{alg:classifier}. All the in-frame classifications are stored in a list $\boldsymbol{\rho}$, to obtain the overall classification of the object. We utilise a moving average filter over $\boldsymbol{\rho}$ with a window of size $w$. After filtering, if any of the elements in the list corresponds to AM, we classify the overall object as articulated. Otherwise, if only RM motion is present in the list, without any AM motion, we consider the object to be rigid. In the end, if only NM motion is present, then the final classification is nondeterministic due to no motion present.

From the algorithm \ref{alg:classifier}, it can be observed that the proposed method is dependent on the type of motion of local regions in ${\bf P}_{i}^{d}$. Hence, if any large noise or background elements are present in ${\bf P}_{i}^{d}$, the registration from these objects will result in no motion (as they are assumed to be fixed). In case of rigid objects, presence of background elements will result in two hash keys, hence classifying it as an articulated object. One way to avoid this issue is to not consider no motion key in the hash map. However, this approach conflicts with the presence of articulated objects in the frame in a way that if some components of an articulated object are static over the whole video sequence, 
this static component will also be avoided, resulting in classifying the object as rigid. To rectify this issue, we masked the intended object during pre-processing, as described in the previous subsection, and only the point cloud information of the intended object is considered in the proposed classification algorithm.
%

%% file: sections/experiments.tex
Experiments are conducted over two benchmark datasets from the literature to demonstrate the ability of the proposed method to classify if an object is articulated or rigid. We evaluate the performance based on the mean accuracy score of classification. Further, qualitative results are presented to showcase the articulation regions detected by the proposed method from a given observation.

\subsection{Dataset Description}
We analyse the performance of the proposed method over RBO~\cite{rbo_dataset} and YCB video dataset~\cite{ycb_video}. RBO dataset provide a set of interaction videos and point clouds for multiple articulated objects. These interaction videos are captured in a controlled environment with motion of objects. For rigid objects, to the best of authors knowledge, no interaction-based datasets with point clouds are available in the literature. However, to showcase the performance of the proposed method for rigid objects, we consider the YCB video dataset which provide multi-object video sequences with point clouds. From RBO dataset, we consider the interaction sequences for $4$ articulated objects, which include, globe, laptop, cabinet, and microwave. Whereas banana, gelatin box, mustard bottle, and power drill are $4$ rigid objects considered from the YCB video dataset for experiments. For each object, at least $10$ video sequences with maximum $200$ frames is utilised for experimentation in this paper.

\subsection{Experimental Setup}
As discussed in Sec.~\ref{sec:method}, the proposed method rely on multiple parameters during the pre-processing and classifier steps. The value of these parameters are decided on multiple trials and statistical properties of the data. For downsampling and smoothening, the value of parameters like $v$ and $r$ is decided by utilising the point cloud properties. Downsample voxel size $v$ is calculated as $\frac{1}{20}$ of the scene object diameter and $r$ is calculated as $5\times v$. Statistical outlier removal parameter $s$ is set to $0.5$ with nearby points threshold as $0.1$ times the number of points in the object point cloud. %
%
Voxel size $x$ in the classifier is set as the $\frac{1}{5}$th of the maximum distance between the points in the point cloud. Other parameters like  $\overline{q}$, $\overline{t}$, and $k$ are decided after multiple trials over a sub-set of the considered data. In this work, $\overline{q} = 0.1$, $\overline{t} = 0.1$, and $k = 5$ are used for the experiments.

\midsepremove
\begin{table}
    \centering
    \caption{Performance Analysis for RBO dataset.}
    \begin{tabular}{c | c}
        \toprule
          Object & Accuracy \% \\
          \midrule
         Laptop & $88.00$ \\
         Globe & $76.00$ \\
         Microwave & $100.00$ \\
         Cabinet & $100.00$ \\
         \midrule
         Average & $91.00$ \\
         \bottomrule
    \end{tabular}
    \label{tab:articulated_results_acc}
\end{table}

\subsection{Performance Analysis}
We analyse the performance of the proposed method both quantitatively and qualitatively. For quantitative analysis, we measure the classification accuracy and average classification probability, while for qualitative analysis, we show the direction and magnitude of registrations between the frames. The per object accuracy is measured as percentage of total number of video sequences correctly classified to total number of video sequences processed for each object. The average classification probability is calculated as the mean probability of each frame set to be classified as AM or RM. Registration direction and magnitude in qualitative results is represented using the arrows with the angle of arrows representing the direction, and the length of arrows representing the magnitude of transformation.

\subsubsection{Accuracy Analysis}
The accuracy performance for both RBO and YCB datasets are present in Table~\ref{tab:articulated_results_acc} and Table~\ref{tab:rigid_results_acc}. From Table~\ref{tab:articulated_results_acc}, it is observed that the proposed method is able to classify Microwave and Cabinet object as articulated with $100\%$ accuracy. For Laptop, $88\%$ accuracy is achieved. A decrease in accuracy is observed due to the presence of mirror screen, which affect the depth sensor and make data very noisy, resulting in registration failure. Further for Globe object, $76\%$ accuracy is achieved, which, considering we are not utilising colour information in registration and the symmetric nature of the object, is significantly good. Overall, the proposed method show an accuracy of $91\%$ to classify the articulated objects in the RBO dataset accurately.

\begin{table}[t]
    \centering
    \caption{Performance Analysis for YCB dataset.}
    \begin{tabular}{c | c }
        \toprule
          Object & Accuracy \%  \\
          \midrule
         Power Drill & $89.47$ \\
         Gelatin Box & $86.36$ \\
         Banana & $95.45$ \\
         Mustard Bottle & $91.30$ \\
         \midrule
         Average & $90.65$ \\
         \bottomrule
    \end{tabular}
    \label{tab:rigid_results_acc}
\end{table}

For rigid object from YCB dataset, the proposed method is able to classify rigid objects with $90.65\%$ accuracy as shown in Table~\ref{tab:rigid_results_acc}. Power Drill, Banana, and Mustard Bottle objects are classified with $89.47\%$, $95.45\%$, and $91.30\%$ respectively. $86.36\%$ accuracy for Gelatin Box is achieved. The lower accuracy is observed for Gelatin Box that is symmetric due to which local registration is impacted. From this analysis, we demonstrate that the proposed method is able to classify an object with high accuracy, without the need of object model, labelled data, and articulation constraints.

\begin{table}
    \centering
    \caption{Per scene sequence classification probabilities for RBO dataset.}
    \begin{tabular}{c | c | c}
        \toprule
          Object & Probability of rigid & { \bf Probability of articulated} \\
          \midrule
         Laptop & $0.1472$ & $0.8528$ \\
         Globe & $0.0041$ & $0.9959$ \\
         Microwave & $0.0251$ & $0.9749$ \\
         Cabinet & $0.0543$ & $0.9457$ \\
         \midrule
         Average & $0.0577$ & $0.9423$ \\
         \bottomrule
    \end{tabular}
    \label{tab:articulated_results_prob}
\end{table}
\begin{table}
    \centering
    \caption{Per scene sequence classification probabilities for YCB dataset.}
    \begin{tabular}{c | c | c }
        \toprule
          Object & {\bf Probability of rigid} & Probability of articulated  \\
          \midrule
         Power Drill & $0.8744$ & $0.1256$ \\
         Gelatin Box & $0.7380$ & $0.2620$ \\
         Banana & $0.9847$ & $0.0153$ \\
         Mustard Bottle & $0.8433$ & $0.1567$ \\
         \midrule
         Average & $0.8601$ & $0.1399$ \\
         \bottomrule
    \end{tabular}
    \label{tab:rigid_results_prob}
\end{table}

\subsubsection{Probability Analysis}
Table~\ref{tab:articulated_results_prob} and~\ref{tab:rigid_results_prob} represents the probability of a set of frames to be classified as articulated or rigid. As discussed in Sec.~\ref{sec:method}, each frame set is classified as AM, RM, or NM, while the final classification is based on the max-count filtered output. In Table~\ref{tab:articulated_results_prob}, performance for objects from RBO dataset is shown and it can be observed that probability for a set of frames to be classified as AM is greater than for RM and NM. Similarly, for rigid objects from YCB dataset, it can be observed from Table~\ref{tab:rigid_results_prob} that the probability for set of frames to be classified as RM is higher than for AM and NM. This analysis shows that the proposed algorithm is identifying the type of objects with higher probabilities even for a single set of frames, hence demonstrating the effectiveness of the proposed method.

\subsubsection{Qualitative analysis}
\begin{figure}
\centering
\includegraphics[width=0.95\columnwidth]{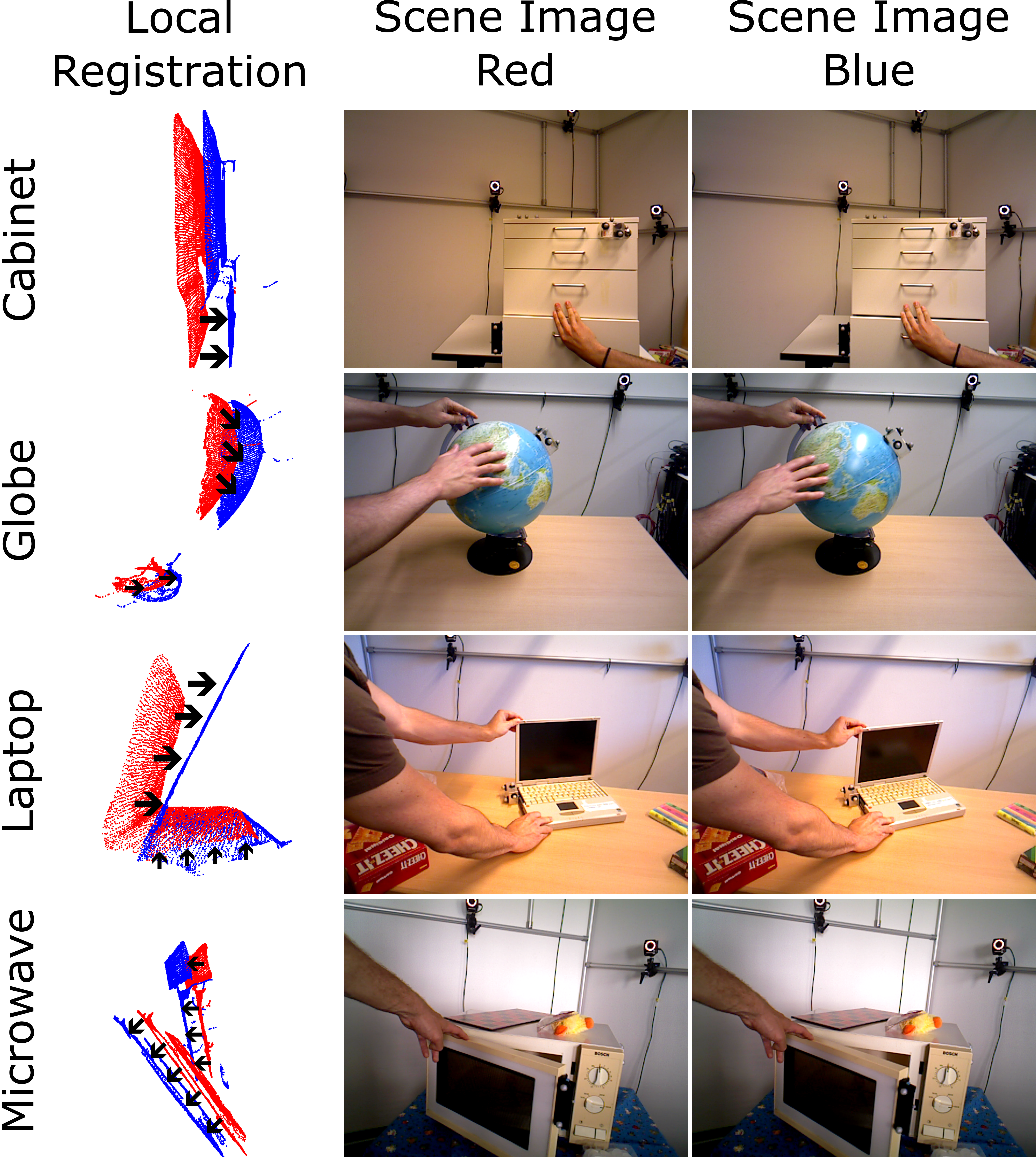}
\caption{Local region registration results for RBO dataset. The arrows represent the direction and magnitude of local region registrations. It is observed that there are multiple unique transformations between two frames for all the objects and hence are classified as articulated.}
\label{fig:rbo_single}
\end{figure}

\begin{figure}
\centering
\includegraphics[width=0.95\columnwidth]{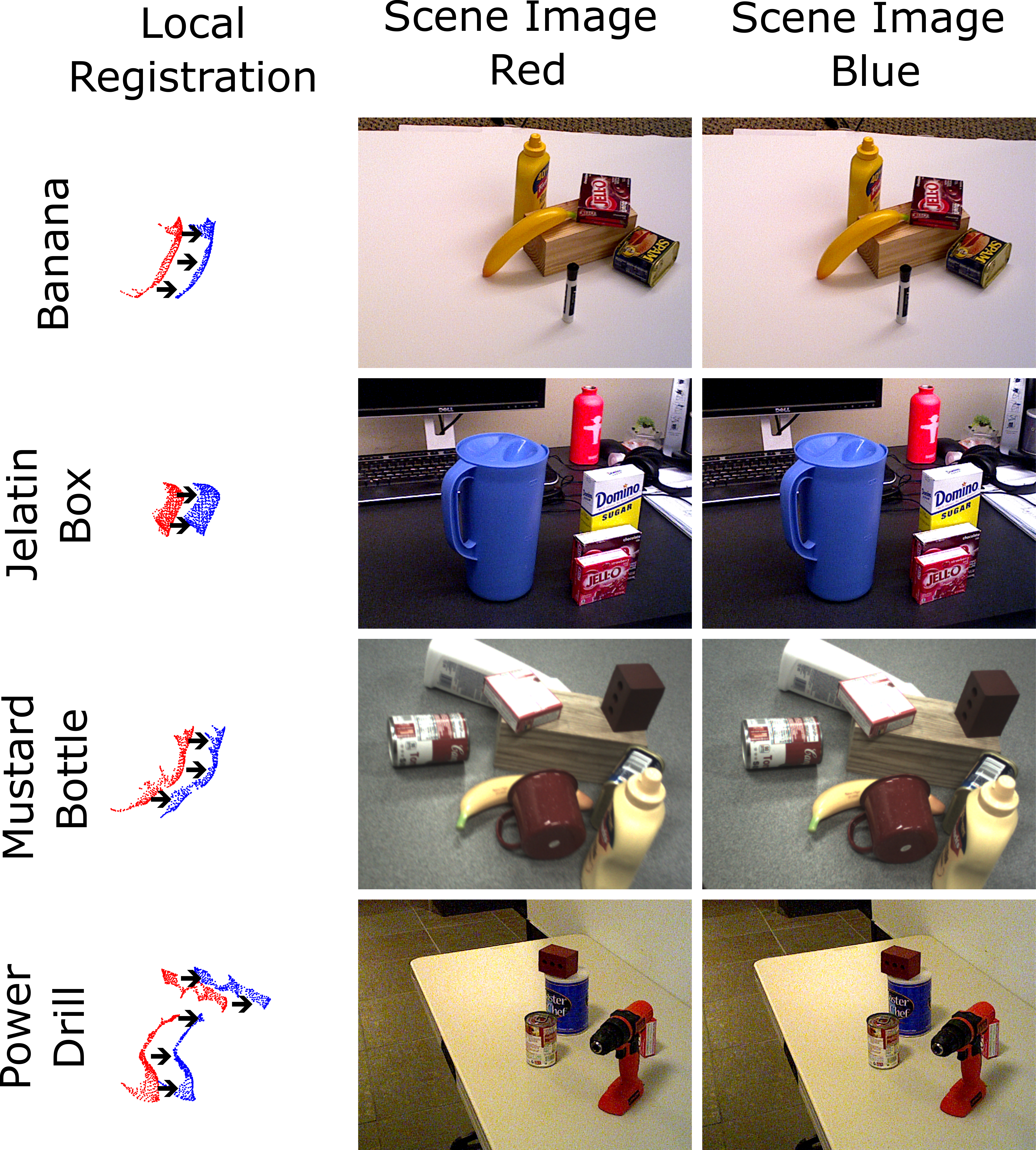}
\caption{Local region registration results for YCB dataset. The arrows represent the direction and magnitude of local region registrations. It is observed that there are multiple unique transformations between two frames for all the objects and hence are classified as rigid.}
\label{fig:ycb_single}
\end{figure}

The local registration results for RBO and YCB datasets, along with the respective rgb images for the two frames are shown in Fig.~\ref{fig:rbo_single} and \ref{fig:ycb_single}. In Fig.~\ref{fig:rbo_single}, it can be observed that more than $1$ unique registration are obtained between the frames and hence they are classified as articulated objects. Similarly, in Fig.~\ref{fig:ycb_single}, it can be observed that a single unique transformation is obtained for all the local regions; hence, classified as rigid objects.

%% file: sections/conclusion.tex
A registration-based local region-to-region mapping approach for articulated object classification has been proposed in this paper. We formulate the object articulation classification as a temporal movement detection method, wherein, we consider the object point cloud from two consecutive time frame observation and identify unique local motions between the two frames. An object is then classified to be articulated if the object is undergoing more than one unique motion during the observations. Otherwise, the object is either in rigid motion or no motion. Due to the model-free nature of the proposed method, it is applicable to wide variety of articulated objects as compared to state-of-the-art methods in the literature. Further, the proposed method has no labelled data requirements. The experimental results with two benchmark datasets demonstrated that the proposed method is able to classify the articulated objects with high accuracy. In future, we plan to extend this work for articulated object modelling and tracking.

%% file: main.bbl
\begin{thebibliography}{10}
\providecommand{\url}[1]{#1}
\csname url@samestyle\endcsname
\providecommand{\newblock}{\relax}
\providecommand{\bibinfo}[2]{#2}
\providecommand{\BIBentrySTDinterwordspacing}{\spaceskip=0pt\relax}
\providecommand{\BIBentryALTinterwordstretchfactor}{4}
\providecommand{\BIBentryALTinterwordspacing}{\spaceskip=\fontdimen2\font plus
\BIBentryALTinterwordstretchfactor\fontdimen3\font minus
  \fontdimen4\font\relax}
\providecommand{\BIBforeignlanguage}[2]{{%
\expandafter\ifx\csname l@#1\endcsname\relax
\typeout{** WARNING: IEEEtran.bst: No hyphenation pattern has been}%
\typeout{** loaded for the language `#1'. Using the pattern for}%
\typeout{** the default language instead.}%
\else
\language=\csname l@#1\endcsname
\fi
#2}}
\providecommand{\BIBdecl}{\relax}
\BIBdecl

\bibitem{marturi2016towards}
N.~Marturi, A.~Rastegarpanah, C.~Takahashi, M.~Adjigble, R.~Stolkin, S.~Zurek,
  M.~Kopicki, M.~Talha, J.~A. Kuo, and Y.~Bekiroglu, ``Towards advanced robotic
  manipulation for nuclear decommissioning: A pilot study on tele-operation and
  autonomy,'' in \emph{2016 International Conference on Robotics and Automation
  for Humanitarian Applications (RAHA)}.\hskip 1em plus 0.5em minus 0.4em\relax
  IEEE, 2016, pp. 1--8.

\bibitem{TowardsUAO}
J.~Sturm, C.~Stachniss, V.~Pradeep, C.~Plagemann, K.~Konolige, and W.~Burgard,
  ``Towards understanding articulated objects,'' in \emph{Proc. of the Workshop
  on Robot Manipulation at Robotics: Science and Systems Conference (RSS)},
  2009.

\bibitem{markerless_motion_capture}
C.-W. Chun, O.~Jenkins, and M.~Mataric, ``Markerless kinematic model and motion
  capture from volume sequences,'' in \emph{2003 IEEE Computer Society
  Conference on Computer Vision and Pattern Recognition, 2003. Proceedings.},
  vol.~2, 2003, pp. II--II.

\bibitem{articulated_free_form_hand}
\BIBentryALTinterwordspacing
Z.~Fan, O.~Taheri, D.~Tzionas, M.~Kocabas, M.~Kaufmann, M.~J. Black, and
  O.~Hilliges, ``Articulated objects in free-form hand interaction,'' 2022.
  [Online]. Available: \url{https://arxiv.org/abs/2204.13662}
\BIBentrySTDinterwordspacing

\bibitem{Understanding_3d_articulation_in_internet_video}
S.~Qian, L.~Jin, C.~Rockwell, S.~Chen, and D.~F. Fouhey, ``Understanding 3d
  object articulation in internet videos,'' in \emph{Proceedings of the
  IEEE/CVF Conference on Computer Vision and Pattern Recognition (CVPR)}, June
  2022, pp. 1599--1609.

\bibitem{visual_articulated_parts_identification}
V.~Zeng, T.~E. Lee, J.~Liang, and O.~Kroemer, ``Visual identification of
  articulated object parts,'' in \emph{2021 IEEE/RSJ International Conference
  on Intelligent Robots and Systems (IROS)}, 2021, pp. 2443--2450.

\bibitem{model_articulated_in_image}
D.~Meyer, J.~Denzler, and H.~Niemann, ``Model based extraction of articulated
  objects in image sequences for gait analysis,'' in \emph{Proceedings of
  International Conference on Image Processing}, vol.~3, 1997, pp. 78--81
  vol.3.

\bibitem{online_recursive_perception}
R.~Martín~Martín and O.~Brock, ``Online interactive perception of articulated
  objects with multi-level recursive estimation based on task-specific
  priors,'' in \emph{2014 IEEE/RSJ International Conference on Intelligent
  Robots and Systems}, 2014, pp. 2494--2501.

\bibitem{sar_images}
\BIBentryALTinterwordspacing
G.~J. III and B.~Bhanu, ``Recognizing articulated objects in sar images,''
  \emph{Pattern Recognition}, vol.~34, no.~2, pp. 469--485, 2001. [Online].
  Available:
  \url{https://www.sciencedirect.com/science/article/pii/S0031320399002186}
\BIBentrySTDinterwordspacing

\bibitem{articulated_tracking_render_parts}
Z.~Pezzementi, S.~Voros, and G.~D. Hager, ``Articulated object tracking by
  rendering consistent appearance parts,'' in \emph{2009 IEEE International
  Conference on Robotics and Automation}, 2009, pp. 3940--3947.

\bibitem{articulated_object_recognition_hough_transform}
A.~Beinglass and H.~Wolfson, ``Articulated object recognition, or: how to
  generalize the generalized hough transform,'' in \emph{Proceedings. 1991 IEEE
  Computer Society Conference on Computer Vision and Pattern Recognition},
  1991, pp. 461--466.

\bibitem{mask_rcnn_v2}
\BIBentryALTinterwordspacing
Y.~Li, S.~Xie, X.~Chen, P.~Dollar, K.~He, and R.~Girshick, ``Benchmarking
  detection transfer learning with vision transformers,'' 2021. [Online].
  Available: \url{https://arxiv.org/abs/2111.11429}
\BIBentrySTDinterwordspacing

\bibitem{coco}
T.-Y. Lin, M.~Maire, S.~Belongie, J.~Hays, P.~Perona, D.~Ramanan,
  P.~Doll{\'a}r, and C.~L. Zitnick, ``Microsoft coco: Common objects in
  context,'' in \emph{Computer Vision -- ECCV 2014}, D.~Fleet, T.~Pajdla,
  B.~Schiele, and T.~Tuytelaars, Eds.\hskip 1em plus 0.5em minus 0.4em\relax
  Cham: Springer International Publishing, 2014, pp. 740--755.

\bibitem{saliency_survey}
\BIBentryALTinterwordspacing
A.~Borji, M.-M. Cheng, Q.~Hou, H.~Jiang, and J.~Li, ``Salient object detection:
  A survey,'' \emph{Computational Visual Media}, vol.~5, no.~2, pp. 117--150,
  Jun 2019. [Online]. Available:
  \url{https://doi.org/10.1007/s41095-019-0149-9}
\BIBentrySTDinterwordspacing

\bibitem{sam}
A.~Kirillov, E.~Mintun, N.~Ravi, H.~Mao, C.~Rolland, L.~Gustafson, T.~Xiao,
  S.~Whitehead, A.~C. Berg, W.-Y. Lo, P.~Doll{\'a}r, and R.~Girshick, ``Segment
  anything,'' \emph{arXiv:2304.02643}, 2023.

\bibitem{freesolo}
X.~Wang, Z.~Yu, S.~D. Mello, J.~Kautz, A.~Anandkumar, C.~Shen, and J.~M.
  Alvarez, ``Freesolo: Learning to segment objects without annotations,'' 2022.

\bibitem{segmentation_without_label}
A.~Voynov, S.~Morozov, and A.~Babenko, ``Object segmentation without labels
  with large-scale generative models,'' in \emph{Proceedings of the 38th
  International Conference on Machine Learning}, vol. 139.\hskip 1em plus 0.5em
  minus 0.4em\relax PMLR, 18--24 Jul 2021, pp. 10\,596--10\,606.

\bibitem{icp}
P.~Besl and N.~D. McKay, ``A method for registration of 3-d shapes,''
  \emph{IEEE Transactions on Pattern Analysis and Machine Intelligence},
  vol.~14, no.~2, pp. 239--256, 1992.

\bibitem{rbo_dataset}
R.~Martín-Martín, C.~Eppner, and O.~Brock, ``The rbo dataset of articulated
  objects and interactions,'' 2018.

\bibitem{ycb_video}
Y.~Xiang, T.~Schmidt, V.~Narayanan, and D.~Fox, ``Posecnn: A convolutional
  neural network for 6d object pose estimation in cluttered scenes,''
  \emph{arXiv preprint arXiv:1711.00199}, 2017.

\end{thebibliography}
